\documentclass[runningheads]{llncs}
\usepackage{times}
\usepackage{epsfig}
\usepackage{graphicx}
\usepackage{amsmath}
\usepackage{amssymb}
\usepackage{comment}
\usepackage{caption}
\usepackage{subfig}
\newcommand{\fatx}{{\mathbf{x}}}
\newcommand{\fatz}{{\mathbf{z}}}
\newcommand{\faty}{{\mathbf{y}}}
\newcommand{\iid}{i.i.d}
\usepackage{color}
\usepackage[width=122mm,left=12mm,paperwidth=146mm,height=193mm,top=12mm,paperheight=217mm]{geometry}
\begin{document}
\pagestyle{headings}
\mainmatter
\def\ECCV16SubNumber{0}  

\title{Variational methods for Conditional Multimodal \\Deep Learning} 

\titlerunning{}

\authorrunning{Gaurav Pandey and Ambedkar Dukkipati}

\author{Gaurav Pandey and Ambedkar Dukkipati}
\institute{Department of Computer Science and Automation\\Indian
  Institute of Science\\Email{\{gp88, ad}@csa.iisc.ernet.in}

\maketitle

\begin{abstract}
In this paper, we address the problem of conditional modality learning, whereby one is interested in generating one modality given the other. While it is straightforward to learn a joint distribution over multiple modalities using a deep multimodal architecture, we observe that such models aren't very effective at conditional generation. Hence, we address the problem by learning conditional distributions between the modalities.  We use variational methods for maximizing the corresponding conditional log-likelihood. The resultant deep model, which we refer to as conditional multimodal autoencoder (CMMA), forces the latent representation obtained from a single modality alone to be `close' to the joint representation obtained from multiple modalities. We use the proposed model to generate faces from attributes. We show that the faces generated from attributes using the proposed model, are qualitatively and quantitatively more representative of the attributes from which they were generated, than those obtained by other deep generative models. We also propose a secondary task, whereby the existing faces are modified by modifying the corresponding attributes. We observe that the 
modifications in face introduced by the proposed model are representative of the corresponding modifications in attributes.
\end{abstract}

\section{Introduction}
The problem of learning from several modalities simultaneously has garnered the attention of several deep learning researchers over the past few years~\cite{ngiam2011multimodal}-\cite{sohn2014improved}. This is primarily because of the wide availability of such data, and the numerous real-world applications where multimodal data is used. For instance, speech may be accompanied with text and the resultant data can be used for training speech-to-text or text-to-speech engines. Even within the same medium, several modalities may exist simultaneously, for instance, the plan and elevation of a 3d object, or multiple translations of a text.

The task of learning from several modalities simultaneously is complicated by the fact that the correlations within a modality are often much stronger than the correlations across modalities. Hence, many multi-modal learning approaches such as~\cite{ngiam2011multimodal}\cite{srivastava2012multimodal} try to capture the cross-modal correlations at an abstract latent feature level rather than at the visible feature level. The assumption is that the latent features are comparatively less correlated than the visible features, and hence, the latent features from different modalities can be concatenated and a single distribution can be learnt for the concatenated latent features~\cite{srivastava2012multimodal}.

An alternative approach to capture the joint distribution, is by modelling the conditional distribution across modalities as done in~\cite{sohn2014improved}, whereby the authors make the simplifying assumption that the joint log-likelihood is maximized when the conditional log-likelihood of each modality given the other modality is maximized. While the assumption is untrue in general, the idea of learning conditional distributions to capture the joint distribution has several advantages. In particular, the conditional distributions are often less complex to model, since conditioning on one modality reduces the possibilities for the other modality. Moreover, if the underlying task is to generate one modality given the other, then learning conditional distributions directly addresses this task. 

Hence, we address the problem of multimodal learning by capturing the conditional distributions. In particular, we use a variational approximation to the joint log-likelihood for training. In this paper, we restrict ourselves to directed graphical models, whereby a latent representation is sampled from one modality (referred to as the conditioning modality) and the other modality (referred to as the generated modality) is then sampled from the latent representation. Hence, the model is referred to as conditional multimodal autoencoder~(CMMA).

\section{Problem Formulation and Proposed Solution}
\begin{figure}
\centering
	\begin{minipage}{.45\textwidth}
  \centering
  \includegraphics[width=.7\linewidth]{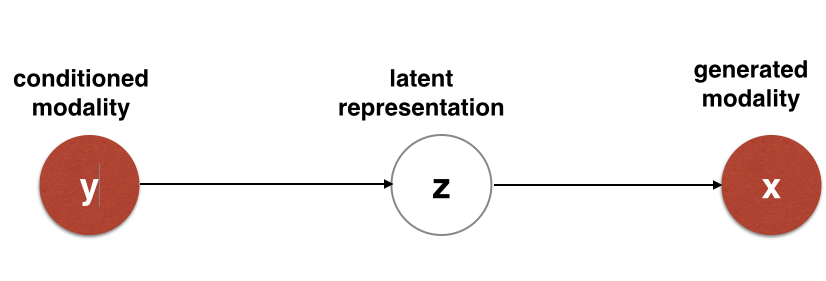}
  \caption{\small A graphical representation of CMMA}
  \label{fig:graphical}
  \end{minipage}
  \hspace{.5cm}
  \begin{minipage}{.45\textwidth}
  \centering
  \includegraphics[width=.7\linewidth]{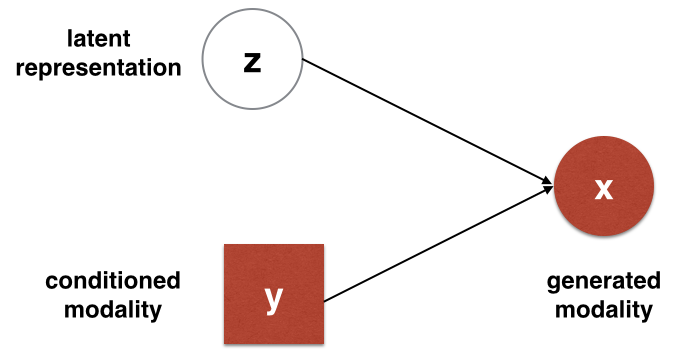}
  \caption{\small A graphical representation of conditional VAE as well as conditional GAN}
  \label{fig:graphical2}
  \end{minipage}
\end{figure}

A formal description of the problem is as follows. We are given an $\iid$ sequence of $N$ datapoints $\{(\fatx^{(1)}, \faty^{(1)}), \dotsc, (\fatx^{(N)}, \faty^{(N)})\}$. For a fixed datapoint $(\fatx, \faty)$, let $\fatx$ be the modality that we wish to generate and $\faty$ be the modality that we wish to condition on. We assume that $\fatx$ is generated by first sampling a real-valued latent representation $\fatz$ from the distribution $p(\fatz|\faty)$, and then sampling $\fatx$ from the distribution $p(\fatx|\fatz)$. The graphical representation of the model is given in Figure~\ref{fig:graphical}. Furthermore, we assume that the conditional distribution of the latent representation $\fatz$ given $\faty$ and the distribution of $\fatx$ given $\fatz$ are parametric. 

Given the above description of the model, our aim is to find the parameters so as maximize the joint log-likelihood of $\fatx$ and $\faty$ for the given sequence of datapoints. The computation of log-likelihood comprises a term for $\log p(\fatx| \faty)$ (referred to as conditional log-likelihood) and a term for $\log p(\faty)$. The computation fo $p(\fatx| \faty)$ requires the marginalization of the latent variable $\fatz$ from the joint distribution $p(\fatx, \fatz| \faty)$.
\begin{align}
p(\fatx|\faty) &= \int p(\fatx, \fatz| \faty) \mathrm{d}\fatz \\
				   &= \int p(\fatx| \fatz)p(\fatz| \faty) \mathrm{d}\fatz	
\end{align}
For most choices of $p(\fatx| \fatz)$ and $p(\fatz| \faty)$, the evaluation of conditional log-likelihood is intractable. Hence, we resort to the minimization of a variational lower bound to the conditional log-likelihood. This is achieved by approximating the posterior distribution of $\fatz$ given $\fatx$ and $\faty$, that is $p(\fatz| \fatx, \faty)$ by a tractable distribution $q(\fatz| \fatx, \faty)$. This is explained in more detail in the following section.

\subsection{The variational bound}~\label{sec:var_bound}
For a given $\iid$ collection of datapoints, $\{(\fatx^{(1)}, \faty^{(1)}), \dotsc, (\fatx^{(N)}, \faty^{(N)})\}$, the log-likelihood can be written as 

\begin{table} 
\centering
\begin{tabular}{|c|c|c|}
\hline
Distribution & Parametric form & Representation\\
\hline
$p(\fatz| \faty)$ & $\mathcal{N}(f_\mu(\faty), e^{f_\sigma(\faty)})$ &  $p_f(\fatz| \faty)$ \\
\hline
$p(\fatx| \fatz)$ & $\mathcal{N}(g_{\mu}(\fatz), e^{g_{\sigma}(\fatz)})$ & $p_g(\fatx| \fatz)$\\
\hline
$q(\fatz|\fatx)$ & $\mathcal{N}(h_{1\mu}(\fatx), e^{h_{1\sigma}(\fatx)})$ & $q_{h_1}(\fatz|\fatx)$\\
\hline
$q(\faty|\fatz)$ & $\mathcal{N}(h_{2\mu}(\fatz), e^{h_{2\sigma}(\fatz)})$ & $q_{h_2}(\faty|\fatz)$\\
\hline
\end{tabular}
\vspace{.3cm}
\caption{Parametric forms for the distributions used in the paper and their representations demonstrating the explicit dependence on $f,g$ and $h$.}
\label{tab:parametric}
\end{table}

\begin{align}
&\log p\left((\fatx^{(1)}, \faty^{(1)}), \dotsc, (\fatx^{(N)}, \faty^{(N)})\right) \notag\\
& \hspace{1cm} = \sum_{i=1}^N\log p(\fatx^{(i)} , \faty^{(i)})
\end{align}

Let the posterior distribution be approximated by a distribution whose graphical representation is shown in Figure~\ref{fig:graphical}.
In particular, $q(\fatz|\fatx)$ be an approximation to the posterior distribution of the latent variables given $\fatx$ and $q(\faty|\fatz)$ be an approximation to the posterior distribution of $\faty$ given $\fatz$ . For an individual datapoint, the conditional log-likelihood can be rewritten as 
\begin{align}
&\log p(\fatx| \faty) \notag\\
& = \mathbb{E}_{q(\fatz| \fatx) } \log \frac{ p(\fatx, \fatz| \faty)}{p(\fatz |\fatx,\faty)} \\
& = \mathrm{KL}\left[q(.|\fatx)   || p(.|\fatx, \faty) \right] + \mathbb{E}_{q(\fatz| \fatx, \faty) } \log \frac{ p(\fatx,  \fatz| \faty)}{ q(\fatz |\fatx)} \notag\\
&\ge \mathbb{E}_{q(\fatz| \fatx) }\log \frac{ p(\fatx,  \fatz| \faty)}{ q(\fatz |\fatx)}\,, ~\label{eq:VariationalLowerBound}
\end{align}
where $\mathrm{KL}(p||q)$ refers to the $\mathrm{KL}$-divergence between the distributions $p$ and $q$ and is always non-negative. Note that the choice of the decomposition of posterior $q(\fatz, \faty| \fatx)$ as $q(\fatz|\fatx)q(\faty|\fatz)$ forces the distribution $q(\fatz|\fatx)$ to be `close' to the true posterior $p(\fatz|\fatx, \faty)$, thereby encouraging the model to learn features from $\fatx$ alone, that are representative about $\faty$ as well.

The term in equation~\eqref{eq:VariationalLowerBound} is referred to as the variational lower bound for the conditional log-likelihood for the datapoint $(\fatx, \faty)$ and will be denoted by $\mathcal{L}(p,q;\fatx, \faty)$. It can further be rewritten as
\begin{align}
&\mathcal{L}(p,q;\fatx, \faty) \notag\\
&=\mathbb{E}_{q(\fatz| \fatx) }\log { p(\fatx|  \fatz)} + \mathbb{E}_{q(\fatz| \fatx) }\log \frac{ p(\fatz| \faty)}{ q(\fatz |\fatx)} \notag\\
& = \mathbb{E}_{q(\fatz| \fatx) }\log { p(\fatx|  \fatz)} - \mathrm{KL}\left[q(\fatz|\fatx) || p(\fatz|\faty)\right] \label{simplifiedLowerBound}
\end{align}

From the last equation, we observe that the variational lower bound can be written as the sum of two terms. The first term is the negative of reconstruction error of $\fatx$, when reconstructed from the encoding $\fatz$ of $\fatx$. The second term ensures that the encoding of $\fatx$ is 'close' to the corresponding encoding of $\faty$, where closeness is defined in terms of $\mathrm{KL}$-divergence between the corresponding distributions.

Adding $\log p(\faty)$ to the above bound, we obtain the lower bound to the joint log-likelihood. It has been shown in~\cite{bengio2013deep} that for learning a distribution from samples, it is sufficient to train the transition operator of a Markov chain, whose stationary distribution is the distribution that we wish to model. Using this idea, we replace $\log p(\faty)$ by $\mathbf{E}_{p(\fatz|\faty)} \log q(\faty| \fatz)$. Note that while the two terms will be quite different, the gradients with respect to the parameters for the two terms is expected to be 'close'.

\subsection{The reparametrization}
In order to simplify the computation of the variational lower bound, we assume that conditioned on $\faty$, the latent representation $\fatz$ is normally distributed with mean $f_\mu(\faty)$ and a diagional covariance matrix whose diagonal entries are given by $e^{f_\sigma(\faty)}$. Moreover, conditioned on $\fatz$, $\fatx$ is normally distributed with mean $g_{\mu}(\fatz)$ and a diagonal covariance matrix whose diagonal entries are given by $e^{g_{\sigma}(\faty)}$. In the rest of the paper, we assume $f_\mu, f_\sigma, g_{\mu}$ and $g_{\sigma}$ to be multi-layer perceptrons. Furthermore, we approximate the posterior distribution of $\fatz$ given $\fatx$ and $\faty$ by a normal distribution with mean $h_\mu(\fatx, \faty)$ and a diagonal covariance matrix whose diagonal entries are given by $e^{h_\sigma(\fatx, \faty)}$, where $h_\mu$ and $h_\sigma$ are again multi-layer perceptrons. In order to make the dependence of the distributions on $f,g$ and $h$ explicit, we represent $p(\fatz|\faty)$ as $p_f(\fatz|\faty)$, $p(\fatx|\fatz)$ as $p_g(\fatx|\fatz)$ and $q(\fatz|\fatx, \faty)$ as $q_h(\fatz|\fatx, \faty)$. For reference, the parametric forms of the likelihood, prior and posterior distributions and their representations demonstrating the explicit dependence on $f,g$ and $h$ are given in Table~\ref{tab:parametric}. 

The above assumptions simplify the calculation of $\mathrm{KL}$-divergence and $\log p(\fatx| \fatz)$. Let $f_j$ denote the $j^{th}$ component of the function $f$ and the size of the latent representation be $J$. After ignoring the constant terms, the $\mathrm{KL}$-divergence term of the variational lower bound can be written as
\begin{align}
&\mathrm{KL}(q_h(\fatz| \fatx, \faty)|| p_f(\fatz| \faty)) = \notag\\
&\frac{1}{2}  \sum_{j=1}^J \bigg[f_{j\sigma}(\faty) - h_{j\sigma}(\fatx, \faty) +  \exp\left(h_{j\sigma}(\fatx, \faty) - f_{j\sigma}(\faty) \right)  \notag\\
&\hspace{3.5cm} + \left. \frac{(h_{j\mu}(\fatx, \faty) - f_{j\mu}(\faty))^2}{\exp(f_{j\sigma}(\faty))}\right] \label{eq:KLDiv}
\end{align}
The negative reconstruction error term in the variational lower bound in ~\eqref{simplifiedLowerBound} can be obtained by generating samples from the posterior distribution of $\fatz$ given $\fatx$ and $\faty$, and then averaging over the negative reconstruction error. For a fixed $\fatz$, the term can be written as
\begin{align}
\log{p_g(\fatx|\fatz)} = -\left[ \sum_{l=1}^m \frac{(g_{l\mu}(\fatz) - \fatx_l  )^2}{2\exp(g_{l\sigma})} + \sum_{l=1}^m \exp(g_{l\sigma}(\fatz)) \right] \label{eq:recon}
\end{align}
The choice of the posterior allows us to sample $\fatz$ as follows:
\begin{align}
\boldsymbol{\epsilon}& \sim \mathcal{N}(0, I) \\
 \fatz & = h_{\mu}(\fatx, \faty) + \boldsymbol{\epsilon} \odot h_{\sigma} (\fatx, \faty)
\end{align} 
where $\odot$ denotes elementwise multiplication. Hence, the negative reconstruction error can alternatively be rewritten as
\begin{equation}
\mathbb{E}_{\boldsymbol{\epsilon \small{\sim} \mathcal{N}(0,I)} }\log { p_g(\fatx|   h_{\mu}(\fatx, \faty) + \boldsymbol{\epsilon} \odot h_{\sigma} (\fatx, \faty))} \,, \label{eq:totalRecon}
\end{equation}
where $\log{p_g(\fatx|.)}$ is as defined in \eqref{eq:recon}

In order to train the model using first-order methods, we need to compute the derivative of the variational lower bound with respect to the parameters of the model. Let $\theta_f, \theta_g$ and $\theta_h$ be the parameters of $\{f_\mu, f_\sigma\}$, $\{g_\mu, g_\sigma\}$ and $\{h_\mu, h_\sigma\}$ respectively. Note that the KL-divergence term in \eqref{eq:KLDiv} depends only on  $\{f_\mu, f_\sigma\}$, and $\{h_\mu, h_\sigma\}$. Its derivatives with respect to $\theta_f$ and $\theta_h$ can be computed via chain rule.

From \eqref{eq:totalRecon}, the derivative of the negative reconstruction error with respect to $\{\theta_g, \theta_h\}$ is given by
\begin{equation}
 \mathbb{E}_{\boldsymbol{\epsilon \small{\sim} \mathcal{N}(0,I)} } \nabla_{\{\theta_g, \theta_h\}}\log { p_g(\fatx|   h_{\mu}(\fatx, \faty) + \boldsymbol{\epsilon} \odot h_{\sigma} (\fatx, \faty))}
\end{equation}
The term inside the expectation can again be evaluated using chain rule. 
\begin{figure*}
\centering
  \centering
  \includegraphics[width=.8\linewidth, height = .35\linewidth]{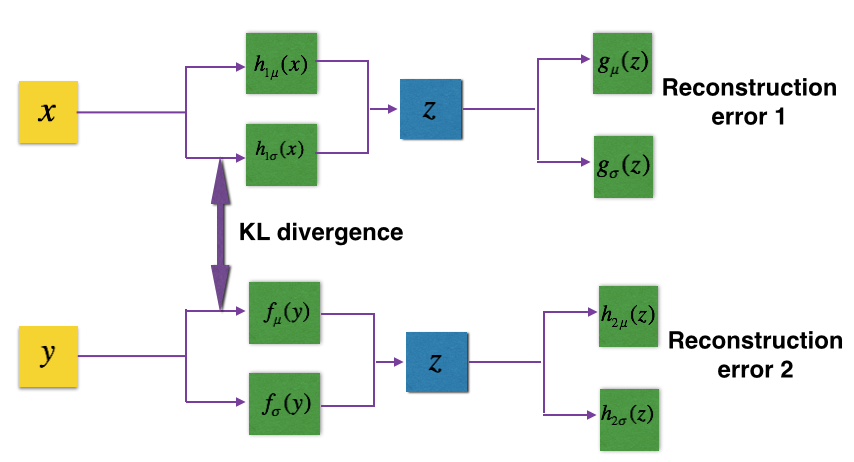}
  \caption{\small A pictorial representation of the implemented model. The KL-divergence between $q_h(\fatz|\fatx, \faty)$ and $p_f(\fatz|\faty)$} is computed using~\eqref{eq:KLDiv} and backpropagated to update the parameters $\theta_h$ and $\theta_f$. Similarly, the negative reconstruction error is computed using equation~\eqref{eq:recon} for the specific $\fatz$ and its gradient is backpropagated to update the parameters $\theta_g$ and $\theta_h$.
  \label{fig:representation}
\end{figure*}%

\subsection{Implementation details}
We use minibatch training to learn the parameters of the model, whereby the gradient of the model with respect to the model parameters $\{\theta_f,\theta_g, \theta_h\}$ is computed for every minibatch and the corresponding parameters updated. While the gradient of the KL-divergence can be computed exactly from~\eqref{eq:KLDiv}, the gradient of the negative reconstruction error in~\eqref{eq:totalRecon} requires one to sample standard normal random vectors, compute the gradient for each sampled vector, and then take the mean. In practise, when the minibatch size is large enough, it is sufficient to sample one standard normal random vector per training example, and then compute the gradient of the negative reconstruction error with respect to the parameters, for this vector. This has also been observed for the case of variational autoencoder in~\cite{kingma2013auto}. 

A pictorial representation of the implemented model is given in Figure~\ref{fig:representation}. Firstly, $\fatx$ and $\faty$ are fed to the neural network $h$ to generate mean and log-variance of the distribution $q_h(\fatz|\fatx, \faty)$. Moreover, $\faty$ is fed to the neural network $f$ to generate the mean and log-variance of the distribution $p_f(\fatz|\faty)$. The KL-divergence between $q_h(\fatz|\fatx, \faty)$ and $p_f(\fatz|\faty)$ is computed using~\eqref{eq:KLDiv}, and its gradient is backpropagated to update the parameters $\theta_f$ and $\theta_h$. Furthermore, the mean and log-variance of $q_h(\fatz|\fatx, \faty)$ are used to sample $\fatz$, which is then forwarded to the neural network $g$ to compute the mean and log-variance of the distribution $p_g(\fatx|\fatz)$. Finally, the negative reconstruction error is computed using equation~\eqref{eq:recon} for the specific $\fatz$ and its gradient is backpropagated to update the parameters $\theta_g$ and $\theta_h$. 

\section{Related Works} \label{relatedWorks}
Over the past few years, several deep generative models have been proposed. They include deep Boltzmann machines (DBM)~\cite{salakhutdinov2009deep}, generative adversarial networks (GAN)~\cite{goodfellow2014generative}, variational autoencoders (VAE)~\cite{kingma2013auto} and generative stochastic networks (GSN)~\cite{bengio2013deep}.

DBMs learn a Markov random field with multiple latent layers, and have been effective in modelling MNIST and NORB data. However, the training of DBMs involves a mean-field approximation step for every instance in the training data, and hence, they are computationally expensive. Moreover, there are no tractable extensions of deep Boltzmann machines for handling spatial equivariance.

All the other models mentioned above, can be trained using backpropagation or its stochastic variant, and hence can incorporate the recent advances in training deep neural networks such as faster libraries and better optimization methods. In particular, GAN learns a distribution on data, by forcing the generator to generate samples that are `indistinguishable' from training data. This is achieved by learning a discriminator whose task is to distinguish between the generated samples and samples in the training data. The generator is then trained to fool the discriminator. Though this approach is intuitive, it requires a careful selection of hyperparameters. Moreover, given the data, one can not sample the latent variables from which it was generated, since the posterior is never learnt by the model.

In a VAE, the posterior distribution of the latent variables conditioned on the data, is approximated by a normal distribution, whose mean and variance are the output of a neural network (distributions other than normal can also be used). This allows approximate estimation of variational log-likelihood which can be optimized using stochastic backpropagation~\cite{icml2014c2_rezende14}.

Both GAN and VAE are directed probabilistic models with an edge from the latent layer to the data. Conditional extensions of both these models for incorporating attributes/labels have also been proposed~\cite{kingma2014semi}\cite{gauthierconditional}\cite{mirza2014conditional}. The graphical representation of a conditional GAN or conditional VAE is shown in Figure~\ref{fig:graphical2}. As can be observed, both these models assume the latent layer to be independent of the attributes/labels. This is in stark contrast with our model CMMA, which assumes that the latent layer is sampled conditioned on the attributes.

It is also informative to compare the variational lower bound of conditional log-likelihood for a CVAE with \eqref{simplifiedLowerBound}. The lower bound for a CVAE is given by
\begin{equation}
\log p(\fatx|\faty) \ge \mathbb{E}_{q(\fatz|\fatx, \faty)} \log p(\fatx| \faty, \fatz) - \mathrm{KL}(q(\fatz|\fatx, \faty) || p(\fatz)) \label{CVAElowerBound}
\end{equation}

Note that while the lower bound in the proposed model CMMA contains a KL-divergence term to explicitly force the latent representation from $\faty$ to be 'close' to the latent representation from both $\fatx$ and $\faty$, there is no such term in the lower bound of CVAE. This proves to be a disadvantage for CVAE as is reflected in the experiments section.

\begin{figure}%
    \centering
    \subfloat[The prior distribution over the latent representations $p(\fatz| \faty)$ for several randomly selected individuals. The width of the circles correspond to standard deviation. Note the high overlap between the priors, despite the attributes being different for the individuals.]{{\includegraphics[width=5cm]{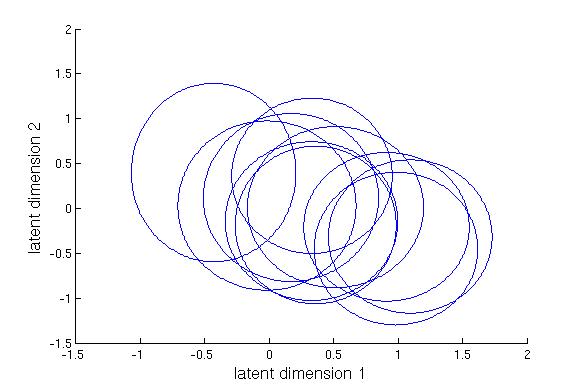} }}%
    \qquad
    \subfloat[The contours of the prior distribution for individual 1 (in blue), and the contours for the posterior distribution for individual 1,2 and 3 (in red). The high uncertainty in the prior causes the posterior distribution of several individuals to lie within the prior distribution of individual 1.]{{\includegraphics[width=5cm]{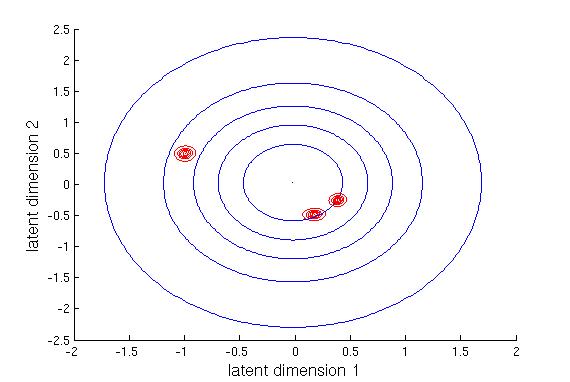} }}%
    \caption{The prior and posterior distributions of latent representation when the dimension of the latent layer if fixed to $2$. }
    \label{prior_and_posterior}
\end{figure}

\section{Experiments}
We consider the task of learning a conditional distribution for the faces given the attributes.
For this task, we use the cropped Labelled Faces in the Wild dataset\footnote{The dataset is available at http://conradsanderson.id.au/lfwcrop/} (LFW)~\cite{LFWTech}, which consists of $13,233$ faces of $5749$ people of which $4069$ people have only one image. The images are of size $64\times 64$ and contain $3$ channels (red green and blue). Of the $13,233$ faces, $13,143$ faces have $73$ attributes associated with them, obtained partially using Amazon Mechanical Turk and partially using attribute classifiers~\cite{kumar2009attribute}. The attributes include `Male', `Asian', `No eye-wear', `Eyeglasses', `Moustache', `Mouth open', `Big nose', `Pointy nose', `Smiling', `Frowning', `Big lips' etc. The data also contains attributes for hair, necklace, earrings etc., though these attributes are not visible in the cropped images. We use the first $10,000$ faces and the corresponding $73$ attributes for training the model, the next $1000$ faces for validation, and keep the remaining faces and their corresponding attributes for testing.

The LFW dataset is very challenging since most people in the dataset have only one image. Moreover, any possible combination of the $73$ attributes occurs at most once in the dataset. This forces the model to learn a mapping from attributes to faces, that is shared across all possible combinations of attributes. In contrast, the face dataset used in~\cite{kulkarni2015deep}, consists of several subsets of faces where only one attribute changes while others remain unchanged. Hence, one can tune the mapping from attributes to faces, one attribute at a time. This, however, isn't possible for LFW.

In order to emphasize this factor, we show the prior ($p(\fatz| \faty)$) and the posterior distribution ($q(\fatz|\fatx, \faty)$) for a 2-dimensional latent representatation of few randomly selected individuals with the modalities $(\fatx, \faty)$ in Figure~\ref{prior_and_posterior}. Note that despite conditioning on the attributes, the prior distributions have high uncertainty, and prior distribution for several attribute combinations $p(\fatz| \faty)$ overlap considerably, particularly in lower dimensions. As the dimensions increase, this overlap decreases however. A VAE, on the other hand, assumes a common prior for all the individuals. Hence, one can think of conditioning in CMMA as tilting the prior of VAE in the direction of the conditioning modality.

 Moreover, the posterior always has much lower variance than the prior. Other than the fact that access to $\fatx$ decreases the uncertainty by a huge amount, the reduced variance is also an artifact of variational methods in general. In particular, for the 2-dimensional latent representations, we observed an average standard deviation of $.04$ for CMMA , and $.036$ for VAE in the posterior distribution of latent representations after $5000$ iterations, which did not reduce further.

\subsection{CMMA architecture}
\begin{figure*}
\centering
  \centering
  \includegraphics[width=.7\linewidth]{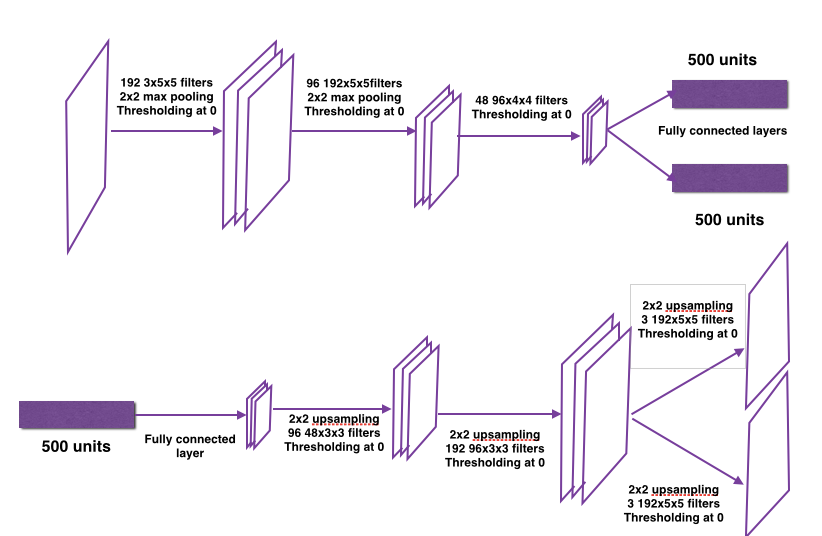}
  \caption{\small The architecture of the MLPs $\{h_\mu, h_\sigma\}$ (top) and $\{g_\mu, g_\sigma\}$ (bottom) of CMMA used in experiments.}
  \label{fig:CMMA}
\end{figure*}

The MLP $\{f_\mu, f_\sigma\}$ of the CMMA used in this paper (refer Figure~\ref{fig:representation}) encodes the attributes, and is a neural network with $800$ hidden units, a soft thresholding unit of the form $\log(1+e^x)$ and two parallel output layers, each comprising of $500$ units. The MLPs $\{h_\mu, h_\sigma\}$  and $\{g_\mu, g_\sigma\}$  are convolution and deconvolution neural networks respectively. The corresponding architectures are given in Figure~\ref{fig:CMMA}.

\subsection{Models used for comparison}
We compare the quantitative and qualitative performance of CMMA against conditional Generative Adversarial Networks~\cite{mirza2014conditional}\cite{gauthierconditional} (CGAN) and conditional Variational Autoencoders~\cite{kingma2013auto} (CVAE). We have tried to ensure that the architecture of the models used for comparison is as close as possible to the architecture of the CMMA used in our experiments. Hence, the generator and discriminator of CGAN and the encoder and decoder of CVAE closely mimic the MLPs $g$ and $h$ of CMMA as described in the previous section.

\subsection{Training}
We coded all the models in Torch~\cite{collobert2011torch7} and trained each of them for $500$ iterations on a Tesla K40 GPU. For each model, the training time was approximately $1$ day. The adagrad optimization algorithm was used~\cite{duchi2011adaptive}. The proposed model CMMA was found to be relatively stable to the selection of initial learning rate, and the variance of the randomly initialized weights in various layers. For CGAN, we selected the learning rate of generator and discriminator and the variance of weights by verifying the conditional log-likelihood on the validation set. Only the results from the best hyperparameters have been reported. We found the CGAN model to be quite unstable to the selection of hyperparameters.

\begin{figure*}
\subfloat[Faces generated from the proposed model CMMA.]{{\includegraphics[width=5.5cm]{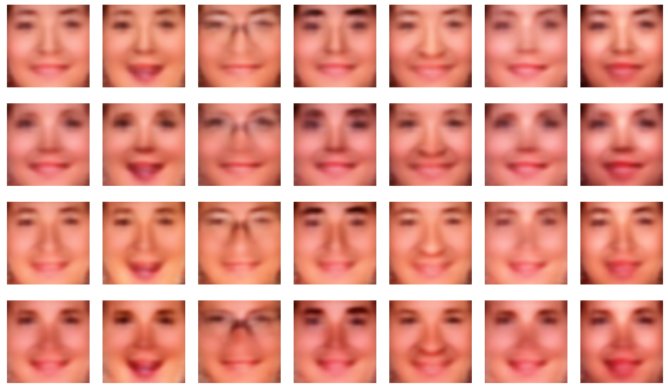} }}
\qquad
\subfloat[Faces generated from CVAE~\cite{kingma2014semi}.]{{\includegraphics[width=5.5cm]{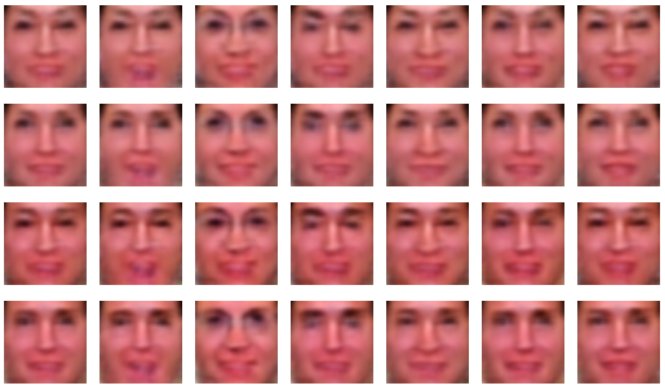} }}
\qquad
\subfloat[Faces generated from CGAN~\cite{gauthierconditional} using the hyperparameters used in~\cite{gauthierconditional}.]{{\includegraphics[width=5.5cm]{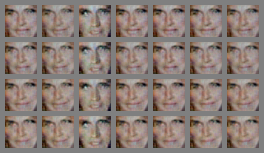} }}
\qquad
\subfloat[Faces generated from CGAN~\cite{gauthierconditional} using the hyperparameters selected by us.]{{\includegraphics[width=5.5cm]{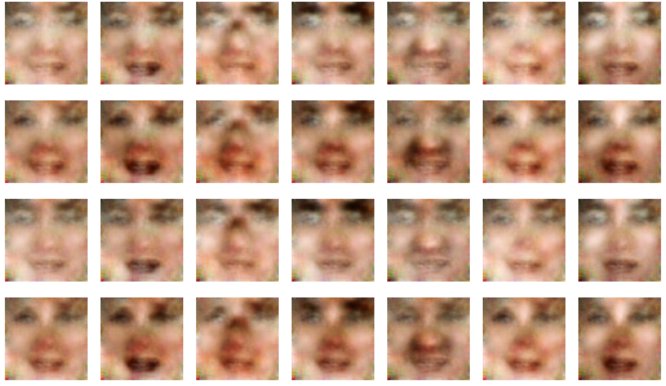} }}
     \caption{\small Faces generated from the attributes using various models  (Best viewed in color). For a fixed model, the $4$ rows correspond to `Female Asian', 'Female Not-Asian', 'Male Asian' and 'Male Not-Asian' in order. The remaining attributes are varied one at a time to generate the $7$ columns. In particular, for each model, the $7$ columns of faces correspond to i) no change, ii) mouth open, iii) spectacles, iv) bushy eyebrows, v) big nose ,vi) pointy nose and vii) thick lips. Note that, for our model CMMA, any change in attributes, such as mouth open, spectacles etc., is clearly reflected in the corresponding face. For other models, this change is not very evident from the corresponding faces.}
 \label{tag_to_face}
\end{figure*}%

\subsection{Quantitative results}
For the first set of experiments, we compare the conditional log-likelihood of the faces given the attributes on the test set for the $3$ models - CMMA, CGAN, and CVAE. A direct evaluation of conditional log-likelihood is infeasible, and for the size of latent layer used in our experiments (500), MCMC estimates of conditional log-likelihood are unreliable. 

For the proposed model CMMA, a variational lower bound to the log-likelihood of the test data can be computed as the difference between the negative reconstruction error and KL-divergence (see \eqref{eq:VariationalLowerBound}). The same can also be done for the CVAE model using \eqref{CVAElowerBound}.

Since we can not obtain the variational lower bound for the other models, we also use Parzen-window based log-likelihood estimation method for comparing the $3$ models. In particular, for a fixed test instance, we condition on the attributes to generate samples from the $3$ models. A Gaussian Parzen window is fit to the generated samples, and the log-probability of the face in the test instance is computed for the obtained Gaussian Parzen window. The $\sigma$-parameter of the Parzen window estimator is obtained via cross-validation on the validation set. The corresponding log-likelihood estimates for the $3$ models are given in Table~\ref{tab:Parzen}.

\begin{table}
\centering
\begin{tabular}{|c|c|c|}
\hline
Model & Conditional Log-likelihood &Variational Lower Bound \\
\hline
\textbf{CMMA} & \textbf{9,487} & \textbf{17,973}\\
CVAE & 8,714 & 14,356\\
CGAN & 8,320 & -\\
\hline
\end{tabular}
\vspace{.3cm}
\caption{\small Parzen window based estimates and variational lower bound to conditional log-likelihood for the test data (Higher means better).}
\label{tab:Parzen}
\end{table}

In both the cases, the proposed model CMMA was able to achieve a better conditional log-likelihood than the other models.

\subsection{Qualitative results}
While the quantitative results do convey a sense of superiority of the proposed model over the other models used in comparison, it is more convincing to look at the actual samples generated by these models. Hence, we compare the three models CGAN, CVAE and CMMA for the task of generating faces from attributes. We also compare the two models CVAE and CMMA for modifying an existing face by changing the attributes. CGAN can not be used for modifying faces because of the uni-directional nature of the model, that is, it is not possible to sample the latent layer from an image in a generative adversarial network. 

\subsubsection{Generating faces from attributes}
\begin{figure}
\centering
  \includegraphics[width=.7\linewidth]{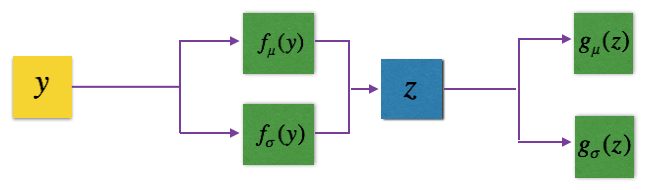}
\caption{\small The model used for generating faces from attributes in CMMA is obtained by removing the MLP $\{h_\mu, h_\sigma\}$ from the CMMA model (refer Figure~\ref{fig:representation}), and connecting the MLP $\{f_\mu, f_\sigma\}$ in its stead.}
  \label{tag_to_face_rep}
\end{figure}

In our first set of experiments, we generate samples from the attributes using the $3$ already trained models.
In a CGAN, the images are generated by feeding noise and attributes to the generator. Similarly, in a CVAE, noise and attributes are fed to the MLP that corresponds to $p(\fatx|\fatz, \faty)$ (see \eqref{CVAElowerBound}) to sample the images. In order to generate images from attributes in a CMMA, we prune the MLP $\{h_\mu, h_\sigma\}$ from the CMMA model (refer Figure~\ref{fig:representation}), and connect the MLP $\{f_\mu, f_\sigma\}$ in its stead as shown in Figure \ref{tag_to_face_rep}.

\begin{figure*} 
\subfloat[Modifying faces with CMMA.]{{\includegraphics[width=.45\linewidth]{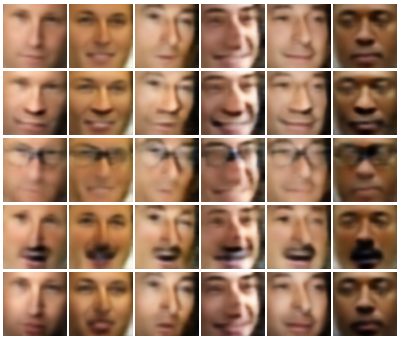} }}
\qquad
\subfloat[Modifying faces with CVAE.]{{\includegraphics[width=.45\linewidth]{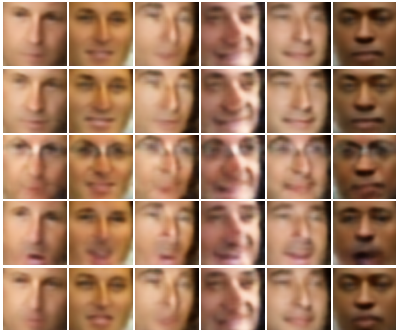} }}
  \caption{\small Modifying the faces in the training data by modifying the corresponding attributes using CMMA and CVAE respectively (Best viewed in color). The rows in each of the above figures correspond to i) No change, ii) Big Nose, iii) Spectacles, iv) Moustache and v) Big lips. Except for spectacles, any other change in attributes is not reflected in the faces modified by CVAE. }
  \label{face_to_face}
\end{figure*}%

\begin{figure}[t]\label{fig:old}
\subfloat[]{\includegraphics[width=.45\linewidth]{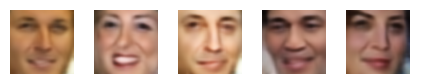}}
\qquad
\subfloat[]{  \includegraphics[width=.45\linewidth]{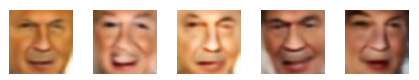}}
   \caption{Modification of faces by setting the old attribute.}
\end{figure}

We set/reset the 'Male' and 'Asian' attributes to generate four possible combinations. The faces are then generated by varying the other attributes one at a time.  
 In order to remove any bias from the selection of images, we set the variance parameter of the noise level to $0$ in CMMA, CVAE and CGAN.  The corresponding faces for our model CMMA, and the other models (CVAE~\cite{kingma2014semi} and CGAN~\cite{gauthierconditional}) are listed in Figure~\ref{tag_to_face}. We have also presented the results from the implementation of CGAN\footnote{https://github.com/hans/adversarial} by the author of~\cite{gauthierconditional}, since the images sampled from CGAN trained by us were quite noisy. 
 
 The $7$ columns of images for each model correspond to the attributes  i) no change, ii) mouth open, iii) spectacles, iv) bushy eyebrows, v) big nose ,vi) pointy nose and vii) thick lips. As is evident from the first image in Figure~\ref{tag_to_face}, CMMA can incorporate any change in attribute such as `open mouth' or `spectacles' in the corresponding face for each of the $4$ rows. However, this does not seem to be the case for the other models. We hypothesize that this is because our model explicitly minimizes the KL-divergence between the latent representation of attributes and the joint representation of face and attributes.

\subsubsection{Varying the attributes in existing faces}
In our next set of experiments, we select a face from the training data, and vary the attributes to generate a modified face. For a CMMA, this can be achieved as follows (also refer Figure~\ref{fig:representation}): 

\begin{enumerate}
\item Let \textit{attr\_orig} be the original attributes of the face and \textit{attr\_new} be the new attributes that we wish the face to possess.
\item Pass the selected face and the \textit{attr\_new} through the MLP $\{h_\mu, h_\sigma\}$. 
\item Pass \textit{attr\_orig} and \textit{attr\_new} through the MLP $\{f_\mu, f_\sigma\}$ and compute the difference.
\item Add the difference to the output of MLP $\{h_\mu, h_\sigma\}$.
\item Pass the resultant sum through the decoder $\{g_\mu, g_\sigma\}$.
\end{enumerate}
As in the previous case, we have the set the variance parameter of noise level to $0$.

Note that, we can not use CGAN for this set of experiments, since, given a face, it is not possible to sample the latent layer in a CGAN. Hence, we only present the results corresponding to our model CMMA and CVAE. The corresponding transformed faces are given in Figure~\ref{face_to_face}. As can be observed, for most of the attributes, our model, CMMA, is successfully able to transform images by removing moustaches, adding spectacles and making the nose bigger or pointy etc.

\subsubsection{Modifying faces with missing attributes}
\begin{figure*} 
\centering
  \subfloat[Unchanged image]{\includegraphics[width=.9\linewidth]{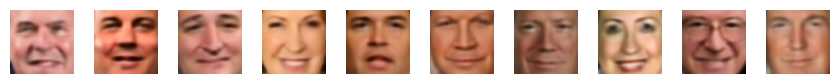}}
  \qquad
  \subfloat[With a big nose] {\includegraphics[width=.9\linewidth]{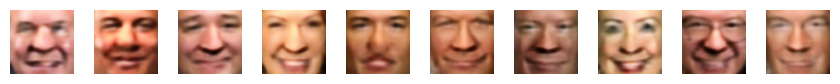}}
  \qquad
 \subfloat[With moustache]{\includegraphics[width=.9\linewidth]{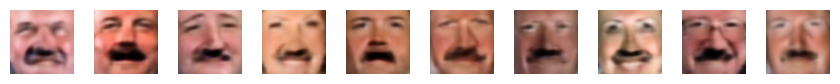}}
 \qquad
 \subfloat[With spectacles and moustache]{\includegraphics[width=.9\linewidth]{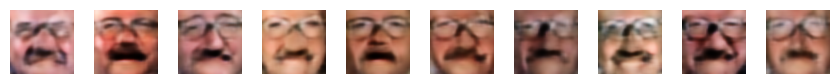}}
\caption{\small Modification of images with missing attributes (Best viewed in color).  }
\label{face_to_face_new}
\end{figure*}

Next, we select a few faces from the web and evaluate the performance of the model for modifying these faces. In order to modify these faces, one needs to sample the attributes conditioned on the faces. The algorithm for modifying the faces mentioned in the previous section is then applied. The corresponding results are given in Figure~\ref{face_to_face_new}.

\section{Concluding Remarks}
In this paper, we proposed a model for conditional modality generation, that forces the latent representation of one modality to be `close' to the joint representation for multiple modalities. We explored the applicability of the model for generating and modifying images using attributes. Quantitative and qualitative results suggest that our model is more suitable for this task than CGAN~\cite{gauthierconditional} and CVAE~\cite{kingma2014semi}. The model proposed, is general and can be used for other tasks such whereby some modalities need to be conditioned whereas others need to be generated, for instance, translation of text or transliteration of speech. We wish to explore the applicability of the model for such problems in future.

\bibliographystyle{splncs}
\bibliography{icml}
\end{document}